\pdfoutput=1

\documentclass[11pt]{article}

\usepackage[preprint]{acl}

\usepackage{times}
\usepackage{latexsym}

\usepackage[T2A,T1]{fontenc}

\usepackage[utf8]{inputenc}

\usepackage[russian,english]{babel}

\def\rub{\begin{otherlanguage*}{russian}}
\def\rue{\end{otherlanguage*}}

\usepackage{microtype}

\usepackage{inconsolata}

\usepackage{graphicx}

\usepackage{todonotes}

\def\todotm#1{\otodo[inline,color=green!40]{#1}}

\def\fix#1{\otodo[inline,color=orange!40]{#1}}

\def\noop#1{}

\let\todotm\noop
\let\fix\noop

\title{Transforming Hidden States into Binary Semantic Features}

\author{Tomáš Musil \and David Mareček\\
Charles University, Faculty of Mathematics and Physics\\
Institute of Formal and Applied Linguistics\\
Prague, Czech Republic \\
  \texttt{\{musil,marecek\}@ufal.mff.cuni.cz} 
  }

\usepackage{tikz}
\usepackage{subcaption}

\usepackage[nonumberlist, nopostdot, hyperfirst=false]{glossaries}
\glsdisablehyper

\newacronym{ica}{ICA}{Independent Component Analysis}
\newacronym{nlp}{NLP}{Natural Language Processing}
\newacronym{nmt}{NMT}{Neural Machine Translation}
\newacronym{pca}{PCA}{Principal Component Analysis}
\newacronym{pos}{POS}{Part of Speech}
\newacronym{llm}{LLM}{Large Language Model}
\newacronym{lm}{LM}{Language Model}

\usepackage{longtable}
\usepackage{array}
\usepackage{booktabs}

\begin{document}
\maketitle
\begin{abstract}
Large language models follow a lineage of many NLP applications that were directly inspired
by distributional semantics, but do not seem to be closely related to it anymore. In this 
paper, we propose to employ the distributional theory of meaning once again. Using
Independent Component Analysis to overcome some of its challenging aspects, we 
show that large language models represent semantic features in their hidden states.
\end{abstract}

\section{Introduction}
\usetikzlibrary{backgrounds}
\usetikzlibrary{shapes}

Distributional semantics have long been a source of inspiration for NLP applications.
However, with the advance of \glspl{llm}, this inspiration has become rather indirect.
In this paper, we show that distributional theories of meaning can still be relevant
in interpreting the hidden states of LLMs and that \gls{ica} can help us overcome
some of the challenges associated with understanding these complex models.

\begin{figure*}

\vspace{-2cm}

\hbox{}\hfill  \begin{tikzpicture}[scale=7,
compnode/.style={circle,fill,draw,blue!40!cyan!15,minimum size=20pt,inner sep=0pt},
innode/.style={regular polygon,regular polygon sides=3,green!40!cyan!10,fill,draw,minimum size=25pt,inner sep=0pt},
tedge/.style={fill=white,rounded corners,opacity=0.9,text opacity=1}]
      \draw[font=\small\bfseries]
        (-0.004, -0.007) node[compnode,label={[align=center]center:Grammar}] (63){}
        (-0.597, -0.806) node[compnode,label={[align=center]center:Components}] (321){}
        (0.958, -0.251) node[compnode,label={[align=center]center:Prepositions}] (276){}
        (0.944, 0.25) node[compnode,label={[align=center]center:Verbs}] (123){}
        (-0.904, -0.436) node[compnode,label={[align=center]center:Actions}] (486){}
        (0.73, -0.68) node[compnode,label={[align=center]center:Verbs}] (164){}
        (-1.0, 0.037) node[compnode,label={[align=center]center:Abstracts}] (235){}
        (0.326, -0.945) node[compnode,label={[align=center]center:Connected}] (292){}
        (-0.864, 0.501) node[compnode,label={[align=center]center:Sciences}] (49){}
        (-0.52, 0.845) node[compnode,label={[align=center]center:Adjectives}] (193){}
        (-0.153, -0.992) node[compnode,label={[align=center]center:Roles}] (99){}
        (-0.004, 0.976) node[compnode,label={[align=center]center:Adverbs}] (154){}
        (0.735, 0.624) node[compnode,label={[align=center]center:Variables}] (298){}
        (0.383, 0.883) node[compnode,label={[align=center]center:Prefixes}] (309){}
        (-0.283, -0.676) node[innode,label={[align=center,font=\scriptsize]center:modifier}] (x321x99){}
        (-0.564, -0.467) node[innode,label={[align=center,font=\scriptsize]center:completion\\formation}] (x321x486){}
        (0.712, -0.002) node[innode,label={[align=center,font=\scriptsize]center:deriving\\emphasizing\\implying}] (x276x123){}
        (0.632, -0.351) node[innode,label={[align=center,font=\scriptsize]center:omitted\\conveyed\\applied}] (x276x164){}
        (-0.685, -0.151) node[innode,label={[align=center,font=\scriptsize]center:exclamations\\capitalizations\\terminations}] (x486x235){}
        (-0.7, 0.2) node[innode,label={[align=center,font=\scriptsize]center:etymologies}] (x235x49){}
        (-0.52, 0.503) node[innode,label={[align=center,font=\scriptsize]center:phonological\\morphological\\etymological}] (x49x193){};
      \begin{scope}[-,align=center,font=\scriptsize, on background layer]
        \draw (63) to node[tedge,pos=0.5] {usage\\form\\forms} (321);
        \draw (63) to node[tedge,pos=0.5] {determiner\\modifier\\intensifier} (99);
        \draw (63) to node[tedge,pos=0.5] {formed\\contracted\\negated} (164);
        \draw (63) to node[tedge,pos=0.5] {combination\\combinations\\combining} (292);
        \draw (63) to node[tedge,pos=0.5] {conjugation\\formation\\derivations} (486);
        \draw (63) to node[tedge,pos=0.5] {search\_word\\all\_words\\count\_word} (298);
        \draw (63) to node[tedge,pos=0.5] {spellings\\pronunciations\\etymologies} (235);
        \draw (63) to node[tedge,pos=0.5] {etymology\\phonology\\morphology} (49);
        \draw (63) to node[tedge,pos=0.5] {used\\with\\derived} (276);
        \draw (63) to node[tedge,pos=0.5] {conj\\ad\\adv} (309);
        \draw (63) to node[tedge,pos=0.5] {grammatically\\semantically\\phonetically} (154);
        \draw (63) to node[tedge,pos=0.5] {lexical\\prepositional\\grammatical} (193);
        \draw (63) to node[tedge,pos=0.5] {modifying\\pronouncing\\forming} (123);
        \draw (321) to node[tedge] {maker\\reader\\trainer} (99);
        \draw (321) to node[tedge] {selection\\preparation\\application} (486);
        \draw (276) to node[tedge] {causing\\using\\appearing} (123);
        \draw (276) to node[tedge] {asked\\looked\\identified} (164);
        \draw[white] (123) to node[] {} (298);
        \draw (486) to node[tedge] {incarcerations\\recognitions\\identifications} (235);
        \draw (164) to node[tedge] {joined\\bridged\\merged} (292);
        \draw (235) to node[tedge] {chemistries\\theologies\\ecologies} (49);
        \draw (292) to node[tedge] {combiner\\unifier\\connecter} (99);
        \draw (49) to node[tedge] {epistemological\\mineralogical\\epidemiological} (193);
        \draw (193) to node[tedge] {rotationally\\institutionally\\structurally} (154);
        \draw[white] (154) to node[] {} (309);
        \draw[white] (298) to node[] {} (309);
      \end{scope}
    \end{tikzpicture}\hfill\hbox{}

\vspace{1cm}

{\small
\centering
\begin{tabular}{p{.12\textwidth}>{\em}p{.327\textwidth} p{.12\textwidth}>{\em}p{.327\textwidth}}
\toprule
Component & Words & Component & Words \\
\midrule
Verbs & modifies, expresses, conveys, connotes, denotes & Transformation & pluralize, italicized, anglicized, latinized, anglicised \\
Identifiers & familienname, name, location.phrasename, surnames, names & MachineLearning & n-gram, bag-of-words, n-grams, ngram, stop\_words \\
Adjectives & metonymic, metaphoric, phonetic, syllabic, syntactic & Literature & proverb, proverbs, idioms, idiom, phrases \\
MathScience & homophones, homophone, conjunctions, prefix, adjectives & Misspellings & begining, refering, pronounciation, defination, defintions \\
Dutch & afkorting, woorden, synoniemen, betekenis, uitspraak & Wikipedia & übersetzung, Übersetzung, aussprache, abkürzung, verwendet \\
\bottomrule
\end{tabular}}
\caption{This is component number 63 from the ICA-transformed hidden states of the Llama 3 70B model,
representing \emph{Grammar}. The outer circle shows the components that share words with this component.
The 10 components that did not fit in the graph are listed in the table bellow the graph (together with top 5 words that combine
the listed component and the central component in the graph). See the caption of Figure~\ref{fig:triangle} for explanation of the
graphic symbols.
}
\label{fig:grammar}
\end{figure*}

\begin{figure}
 \hbox{}\hfill  \begin{tikzpicture}[scale=1.75,
compnode/.style={circle,fill,draw,blue!40!cyan!15,minimum size=30pt,inner sep=0pt},
innode/.style={regular polygon,regular polygon sides=3,green!40!cyan!10,fill,draw,minimum size=50pt,inner sep=0pt},
tedge/.style={fill=white,rounded corners,opacity=0.9,text opacity=1}]
      \draw[font=\normalsize\bfseries]
        (-2, 0) node[compnode,label={[align=center]center:MachineLearning}] (482){}
        (1, -1.5) node[compnode,label={[align=center]center:Roles}] (99){}
        (1, 1.5) node[compnode,label={[align=center]center:Grammar}] (63){}
        (-0.25, 0.0) node[innode,label={[align=center,font=\footnotesize]center:tokenizer\\bertwordpiecetokenizer\\snowballstemmer}] (x99x63){};
      \begin{scope}[-,align=center,font=\footnotesize, on background layer]
        \draw (482) to node[tedge,pos=0.5] {recommender\\annotator\\labelbinarizer} (99);
        \draw (482) to node[tedge,pos=0.5] {n-gram\\bag-of-words\\n-grams} (63);
        \draw (99) to node[tedge,pos=0.55] {determiner\\modifier\\intensifier} (63);
      \end{scope}
    \end{tikzpicture}\hfill\hbox{}
 
  \caption{Combining components. The blue circle nodes represent the components, the edges represent the connections between them. The labels on the edges show the words that are shared between the components. The words on the triangle in the middle belong to all three components.}
 \label{fig:triangle}
 \end{figure}
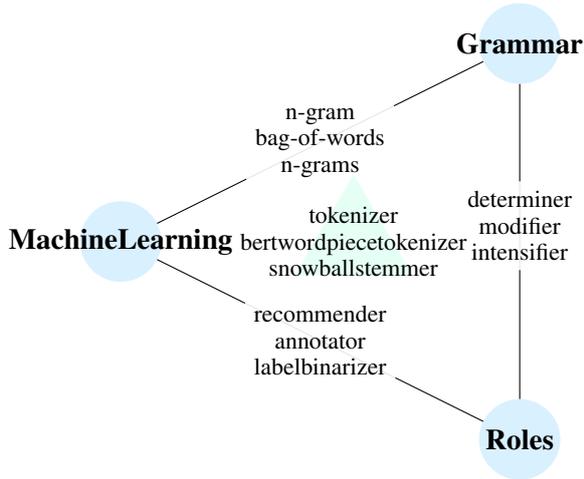

We propose to interpret dimensions of a hidden state of an LLM as linear combinations of
values of a binary vector. We show that it is possible to use \gls{ica} to
transform the hidden states of the model into binary vectors and that they are
interpretable as semantic features. Furthermore, we show that these features are
compositional. An example of three such semantic features (or components) is
shown in Figure~\ref{fig:triangle}.

It has been previously demonstrated (see Sec.~\ref{sec:related}) that ICA can
produce semantic features on word embeddings. The main contributions of this
paper are: showing that it works also on the hidden states of LLMs and showing
that the components can be combined.

\section{Background}

In this section, we explain \gls{ica}, discuss distributional theories of
meaning and review related work.

\subsection{Independent Component Analysis}

\Gls{ica} \cite{comon_independent_1994} is an algorithm originally developed for
separating sources of sound in audio recordings. The model assumes that we have
$n$ recordings, each one a different linear combination of $n$ sources of signal.
The goal is to reconstruct the original signals from the mixed recordings.

The \gls{ica} algorithm \citep{hyvarinen_independent_2000} consists of:
\begin{enumerate}
    \item optional dimension reduction, usually with \gls{pca},
    \item centering the data (setting the mean to zero) and \emph{whitening} them (setting variance of each component to 1),
    \item iteratively finding directions in the data that are the most non-Gaussian.
\end{enumerate}

The last step is based on the assumption of the central limit theorem: the
mixed signal is a sum of independent variables, therefore it should be closer
to the normal distribution than the variables themselves.

The \gls{ica} algorithm is stochastic; every run gives a slightly different
result. It always returns as many components as we specify before running it
(up to the dimension of the original data). If the data was generated by a
lower number of independent components and some random noise, \gls{ica} will
return some components containing only the noise.
Due to the random initialization, the sign of each resulting component is arbitrary.

When applied on word embeddings, ICA components are almost always one-sided
\citep{musil_exploring_2024}, meaning that only one direction
(positive/negative) is interpretable. For a positive component, this means that
a high positive value indicates the presence of the feature, values near zero or
negative indicate its absence. For a negative component, the opposite is true.


\subsection{Distributional Theories of Meaning}

Distributional theories of meaning are based on the hypothesis,
often atributed to \citet{Harris_1954}, that expressions with similar
meaning will have similar distributions across corpora. From count-based
vector representations, word embedding methods like word2vec, early neural \glspl{lm} to contextual
embeddings in \glspl{llm}, this idea seems to have profoundly influenced the development of NLP.
Therefore, it seems to be a good candidate to explain the success of LLMs.

Distributional theories of meaning offer several advantages
\citep{grindrod_distributional_2023}: they have been succesfully employed in NLP applications, the representations can be extracted from language sources using automated methods, they allow for straightforward measurement of similarity. 
However, these approaches also present certain challenges, the most 
significant being their lack of interpretability, compositionality and granularity of meaning.
In Section~\ref{sec:themodel}, we show how ICA helps to overcome these challenges. 

\subsection{Related Work}
\label{sec:related}

\citet{yamagiwa_discovering_2023} show that ICA can unveil semantic structure
within embeddings of words and images that is consistent across languages and
modalities. \citet{li_exploring_2024} confirmed consistency of ICA within and
across languages.

\citet{yamagiwa_revisiting_2024} discuss interpretation of cosine similarity on
embeddings and show that ICA-transformed embeddings exhibit sparsity, thereby
enhancing interpretability by delineating clear semantic contributions. They also
show that ICA components can be ordered and grouped by their
semantic content \citep{yamagiwa_axis_2024}.

\citet{musil_exploring_2024} use word intruder test to show that ICA components
of word embeddings are interpretable. They provide examples of combinations of
components obtained from word2vec embeddings.

\section{The Proposed Model}
\label{sec:themodel}

To address the issue of interpretability of word embeddings, we propose to model meaning of a word
as consisting of independent binary semantic features.
A word embedding would then
be a vector of real (or floating-point) numbers, where each dimension
represents a linear combination of the semantic features. \gls{ica} would be an algorithm
that finds transformation from the embeddings to the semantic feature vectors.
Transforming the embeddings into vectors that represent semantic
features helps with all three problems of distributional semantics:

Interpretability: if we are able to associate each component of a binary vector
with a semantic feature, we can interpret the values of the vector as indicating
the presence of the corresponding features.

Compositionality: while this approach does not address the issue with asymmetry
of compositionality within language (e.g. ``the cat licks the dog'' versus ``the
dog licks the cat''), for the cases where compositionality does work, it is
straigthforward. Composing two concepts represented by binary vectors of
semantic features by taking their union (to find what object falls under both
concept) or intersection (to find what do the two concepts have in common) has
more intuitive interpretation than adding or multiplying word embeddings.

Granularity of meaning: unlike typical word embeddings, where a singular
dimension has no interpretation, each component of the binary vector
corresponds to a specific semantic feature, which can be interpreted
and manipulated independently.

\section{Demonstration of the Model}

We use the Llama 3 8B and 70B model \citep{dubey_llama_2024}. We are
interested in lexical semantics of words, therefore we need a vocabulary to
analyze. Experiments with various corpora showed us that the choice of
vocabulary affects the results. Because we want to be able to interpret what the
LLM has learned by itself, we need to obtain the vocabulary from the model
itself. Subsequently, we extract representations from the model for each
word in the vocabulary, run ICA on them and binarize the result.


\subsection{Vocabulary}

To generate a vocabulary, we sample from the model\footnote{empty prompt, temperature=1.1,
max\_length=100, top\_k=0} to obtain text sequences. The sampled sequences
contain mostly English text and the distribution of topics seems to be
proportional to the amounts of various document domains used in the training
data \citep{dubey_llama_2024}. We split the generated texts into words using
NLTK \citep{bird_natural_2009}. We repeat this procedure until we have the
desired number of words (in our case 250\,000) that have been seen at least 5
times in the data. We had to generate aproximatelly 94M words to obtain this
size of vocabulary. 

\subsection{Hidden states}

For a given LLM, vocabulary $V$ and number of layer $L$, we obtain the representations 
in the following way: for each word $w \in V$, we run the model with the prompt ``The
meaning of the word <$w$>'' and extract the hidden state vector at the layer 8 at the
last token (following \citet{meng_locating_2022, limisiewicz_debiasing_2023}, who
found that the last token is where most of the information about a word is
accumulated).  This way, we obtain a 4096 (8B model) or 8192 (70B model)
dimensional vector for each word in the vocabulary.

\todotm{zmínit prázdný prompt?}

\subsection{ICA Transformation and Binarization}

On these vectors, we first run PCA to get a more manageable number of dimensions
(512 or 1024). Then we run \gls{ica} (we are using the scikit-learn
\citep{pedregosa_scikit-learn_2011} implementation of the FastICA algorithm
\citep{hyvarinen_fast_1999}) and obtain the same number of components for each
word in the vocabulary. Because the resulting vectors vary significantly in their
norm, we normalize them before applying a threshold to binarize them.

Let $M_{i,c}$ be the value of the component $c$ for the word $i$. The vector of components for the word $i$ will be denoted by $M_i$. We define the normalized matrix $N$ as $$N_{i,c} = \frac{M_{i,c}}{|M_i|}$$
To obtain binary features from the real-valued
components, we use the following formula for each word $i$ in the vocabulary and component $c$:
$$ B_{i,c} = \begin{cases}
	1,& \text{if } |N_{i,c}| > t\\
	0,& \text{otherwise,}
\end{cases}$$
where $t$ is a treshold parameter.

\todotm{how to set the treshold}

\subsection{Presenting the Combinations}

To obtain names for the components, we use GPT4o through the OpenAI API, supply
it with 30 words from one end of the component (from the normalized matrix $N$)
and ask it to give the group of words a name.\footnote{See
Appendix~\ref{app:gptprompt} for the the prompt we used.} To make the
presentation of the results more readable, we ask GPT to only use one word when
naming the component. This can sometimes lead to overly general names (e.g.
``Names'' instead of ``Female Names'' or ``Indian Names''). However, for the
purposes of presenting the relations between components in a legible form, we
find the short name more useful even if we lose a degree of specificity.

We obtain names for both positive and negative end of each component. We assume
that all of the components are uni-directional.  For each component, we look up
the words that have $1$ for that component in the binary matrix $B$ and count
the signs of the corresponding entries in the component matrix $M$. If most of
them are positive, we use the name for the positive end, if most of them are
negative, we use the negative one. 

To show that the components can be combined as semantic features, we construct
a graph, where each node is a component. Two nodes are connected by an edge if
there is more than $k$ words in the vocabulary, that have a $1$ in the matrix
$B$ for both corresponding components. 

\subsection{Results}

\begin{figure*}

\vspace{0cm}

 \hbox{}\hfill  \begin{tikzpicture}[scale=7,
compnode/.style={circle,fill,draw,blue!40!cyan!15,minimum size=20pt,inner sep=0pt},
innode/.style={regular polygon,regular polygon sides=3,green!40!cyan!10,fill,draw,minimum size=25pt,inner sep=0pt},
tedge/.style={fill=white,rounded corners,opacity=0.9,text opacity=1}]
      \draw[font=\small\bfseries]
        (-0.009, 0.001) node[compnode,label={[align=center]center:Instruments}] (37){}
        (-0.892, -0.435) node[compnode,label={[align=center]center:Specialists}] (510){}
        (0.727, 0.627) node[compnode,label={[align=center]center:Writers}] (90){}
        (0.934, 0.246) node[compnode,label={[align=center]center:Tools}] (239){}
        (0.374, 0.889) node[compnode,label={[align=center]center:Literature}] (175){}
        (-0.872, 0.489) node[compnode,label={[align=center]center:Groups}] (291){}
        (-1.0, 0.02) node[compnode,label={[align=center]center:Clubs}] (256){}
        (-0.532, 0.84) node[compnode,label={[align=center]center:French}] (120){}
        (-0.02, 0.978) node[compnode,label={[align=center]center:Abstracts}] (235){}
        (0.925, -0.282) node[compnode,label={[align=center]center:Traditions}] (394){}
        (-0.566, -0.82) node[compnode,label={[align=center]center:Pairs}] (209){}
        (0.704, -0.656) node[compnode,label={[align=center]center:Verbs}] (164){}
        (-0.064, -0.975) node[compnode,label={[align=center]center:Musicians}] (122){}
        (0.339, -0.903) node[compnode,label={[align=center]center:Adjectives}] (138){}
        (-0.549, -0.478) node[innode,label={[align=center,font=\scriptsize]center:bassist/vocalist\\guitarist/vocalist\\singer/guitarist}] (x510x209){}
        (0.695, -0.013) node[innode,label={[align=center,font=\scriptsize]center:didgeridoo\\handpan\\balalaika}] (x239x394){}
        (-0.704, 0.191) node[innode,label={[align=center,font=\scriptsize]center:dancers}] (x291x256){}
        (-0.219, 0.682) node[innode,label={[align=center,font=\scriptsize]center:musiques}] (x120x235){}
        (-0.246, -0.673) node[innode,label={[align=center,font=\scriptsize]center:pop-rock\\country-pop\\folk-pop}] (x209x122){};
      \begin{scope}[-,align=center,font=\scriptsize, on background layer]
        \draw (37) to node[tedge,pos=0.5] {music\\\_music\\piano} (256);
        \draw (37) to node[tedge,pos=0.5] {saxophonists\\musicians\\pianists} (291);
        \draw (37) to node[tedge,pos=0.5] {strummed\\serenaded\\trumpeted} (164);
        \draw (37) to node[tedge,pos=0.55] {indie-rock\\-yhtye\\country-pop} (122);
        \draw (37) to node[tedge,pos=0.5] {cumbia\\shamisen\\dhol} (394);
        \draw (37) to node[tedge,pos=0.5] {musiques\\musics\\músicas} (235);
        \draw (37) to node[tedge,pos=0.66] {melodic\\symphonic\\polyphonic} (138);
        \draw (37) to node[tedge,pos=0.5] {mandolin\\mandolins\\ukulele} (239);
        \draw (37) to node[tedge,pos=0.55] {concerto\\ballad\\cantata} (175);
        \draw (37) to node[tedge,pos=0.4] {actress/singer\\pop/rock\\singer-guitarist} (209);
        \draw (37) to node[tedge,pos=0.5] {guitare\\musique\\chanteur} (120);
        \draw (37) to node[tedge,pos=0.5] {shostakovich\\gershwin\\bartók} (90);
        \draw (37) to node[tedge,pos=0.6] {saxophonist\\pianist\\organist} (510);
        \draw (510) to node[tedge] {dentist\\physician\\poet} (256);
        \draw (510) to node[tedge] {writer/artist\\architect/designer\\author/artist} (209);
        \draw[white] (90) to node[] {} (239);
        \draw[white] (90) to node[] {} (175);
        \draw (239) to node[tedge] {shamisen\\shakuhachi\\handpan} (394);
        \draw[white] (175) to node[] {} (235);
        \draw (291) to node[tedge] {farmers\\soldiers\\lawyers} (256);
        \draw (291) to node[tedge] {joueurs\\vainqueurs\\étudiants} (120);
        \draw (120) to node[tedge] {musiques\\cléschronologies\\plastiques} (235);
        \draw[white] (394) to node[] {} (164);
        \draw (209) to node[tedge] {rock/metal\\pop/rock\\ac-dc} (122);
        \draw[white] (164) to node[] {} (138);
        \draw[white] (122) to node[] {} (138);
      \end{scope}
    \end{tikzpicture}\hfill\hbox{}

\vspace{1cm}
    
 {\small
\centering
\begin{tabular}{p{.12\textwidth}>{\em}p{.327\textwidth} p{.12\textwidth}>{\em}p{.327\textwidth}}
\toprule
Component & Words & Component & Words \\
\midrule
Accents & opéra, bodhrán, música, canción, trío & Wikipedia & klavier, musikalische, konzerte, jazzmusiker, schlagzeug \\
Components & music, songs, song, instrument, instruments & Verbs & rehearsing, orchestrating, serenading, improvising, strumming \\
Connected & harmony, harmonies, duets, unison, duet & Italian & musiche, gesualdo, cantabile, giacchino, paganini \\
Nonsense & raag, bassoon, koor, moog, veena & Sounds & strumming, jangle, strum, strums, fiddling \\
Russian & shostakovich, prokofiev, balalaika, mussorgsky, glazunov & Documentos & concierto, guitarra, orquesta, canciones, sonido \\
Suffixes & philharmonia, melodia, symphonia, sinfonia, harmonia & \rub Информатика \rue & \rub Композитор, Музыкант, Орган, Песня, песня \rue \\
Adjectives & orchestral, choral, musical, symphonic, instrumental & Adverbs & musically, vocally, harmonically, rhythmically, acoustically \\
Film & operatic, opera, operas, musical, opéra & Roles & synthesizer, sequencer, arranger, composer, improviser \\
Numbers & quintet, quartet, septet, quartets, trio & Multi & multi-instrument, multitrack, multi-track, multi-instrumentalist, polyphonic \\
\bottomrule
\end{tabular}}

  \caption{This is component number 37 from the ICA-transformed hidden states of the Llama 3 70B model, representing (musical) \emph{Instruments}. The outer circle shows the components that share words with this component. The 18 components that did not fit in the graph are listed in the table bellow the graph (together with top 5 words that combine the listed component and the central component in the graph). See the caption of Figure~\ref{fig:triangle} for explanation of the graphic symbols.}
 \label{fig:instruments}
 \end{figure*}
\begin{figure*}
 \hbox{}\hfill  \begin{tikzpicture}[scale=7.5,
compnode/.style={circle,fill,draw,blue!40!cyan!15,minimum size=20pt,inner sep=0pt},
innode/.style={regular polygon,regular polygon sides=3,green!40!cyan!10,fill,draw,minimum size=25pt,inner sep=0pt}]
      \draw[font=\small\bfseries]
        (-0.008, 0.015) node[compnode,label={[align=center]center:Pharmaceuticals}] (64){}
        (0.191, -0.914) node[compnode,label={[align=center]center:Anti}] (312){}
        (-0.319, -0.89) node[compnode,label={[align=center]center:Participles}] (469){}
        (-0.469, 0.891) node[compnode,label={[align=center]center:Adjectives}] (138){}
        (-0.987, -0.179) node[compnode,label={[align=center]center:Medications}] (191){}
        (-0.74, -0.623) node[compnode,label={[align=center]center:Drugs}] (180){}
        (0.623, 0.77) node[compnode,label={[align=center]center:Cells}] (210){}
        (0.646, -0.674) node[compnode,label={[align=center]center:Dermatology}] (476){}
        (0.094, 1.0) node[compnode,label={[align=center]center:Nutrients}] (227){}
        (0.932, 0.257) node[compnode,label={[align=center]center:Biotech}] (322){}
        (0.915, -0.239) node[compnode,label={[align=center]center:Roles}] (99){}
        (-0.877, 0.487) node[compnode,label={[align=center]center:Psychopathology}] (391){}
        (-0.05, -0.673) node[innode,label={[align=center,font=\scriptsize]center:antiperspirant}] (x312x469){}
        (0.312, -0.592) node[innode,label={[align=center,font=\scriptsize]center:antiperspirant}] (x312x476){}
        (-0.399, -0.564) node[innode,label={[align=center,font=\scriptsize]center:inhalants}] (x469x180){}
        (-0.507, 0.521) node[innode,label={[align=center,font=\scriptsize]center:neuroleptic}] (x138x391){}
        (-0.65, -0.26) node[innode,label={[align=center,font=\scriptsize]center:painkillers}] (x191x180){}
        (-0.701, 0.157) node[innode,label={[align=center,font=\scriptsize]center:neuroleptic\\neuroleptics\\antipsychotic}] (x191x391){}
        (0.583, -0.339) node[innode,label={[align=center,font=\scriptsize]center:cleansers\\moisturizer}] (x476x99){};
      \begin{scope}[-,align=center,font=\scriptsize, on background layer]
        \draw (64) to node[fill=white,rounded corners,pos=0.5] {biopesticides\\neuroprotective\\bioactives} (322);
        \draw (64) to node[fill=white,rounded corners,pos=0.5] {antioxidant\\antioxidants\\phytoestrogens} (227);
        \draw (64) to node[fill=white,rounded corners,pos=0.5] {emulsifier\\cleanser\\oxidizers} (99);
        \draw (64) to node[fill=white,rounded corners,pos=0.5] {psychotropic\\antipsychotic\\neuroleptics} (391);
        \draw (64) to node[fill=white,rounded corners,pos=0.5] {anthelmintic\\diuretic\\anxiolytic} (138);
        \draw (64) to node[fill=white,rounded corners,pos=0.5] {chemokine\\procoagulant\\cytotoxin} (210);
        \draw (64) to node[fill=white,rounded corners,pos=0.5] {psychostimulant\\hallucinogen\\hallucinogens} (180);
        \draw (64) to node[fill=white,rounded corners,pos=0.55] {relaxant\\sclerosant\\retardants} (469);
        \draw (64) to node[fill=white,rounded corners,pos=0.5] {anticonvulsant\\anti-coagulant\\anti-corrosive} (312);
        \draw (64) to node[fill=white,rounded corners,pos=0.5] {emollient\\moisturizers\\exfoliants} (476);
        \draw (64) to node[fill=white,rounded corners,pos=0.5] {decongestants\\anticonvulsants\\antidepressants} (191);
        \draw[white] (312) to node[] {} (469);
        \draw[white] (312) to node[] {} (476);
        \draw[white] (469) to node[] {} (180);
        \draw (138) to node[fill=white,rounded corners] {pantothenic\\lipoic\\ascorbic} (227);
        \draw (138) to node[fill=white,rounded corners] {schizophrenic\\hebephrenic\\mesolimbic} (391);
        \draw (191) to node[fill=white,rounded corners] {phencyclidine\\carisoprodol\\butorphanol} (180);
        \draw (191) to node[fill=white,rounded corners] {olanzapine\\clomipramine\\aripiprazole} (391);
        \draw (210) to node[fill=white,rounded corners] {microvessels\\biomarker\\mechanotransduction} (322);
        \draw (210) to node[fill=white,rounded corners] {protein\\carnitine\\collagen} (227);
        \draw[white] (476) to node[] {} (99);
        \draw[white] (322) to node[] {} (99);
      \end{scope}
    \end{tikzpicture}\hfill\hbox{}
  \caption{This is component number 64 from the ICA-transformed hidden states of the Llama 3 70B model,
representing \emph{Pharmaceutics}. The outer circle shows the components that share words with this component. See the caption of Figure~\ref{fig:triangle} for explanation of the
graphic symbols.}
 \label{fig:pharmaceuticals}
 \end{figure*}
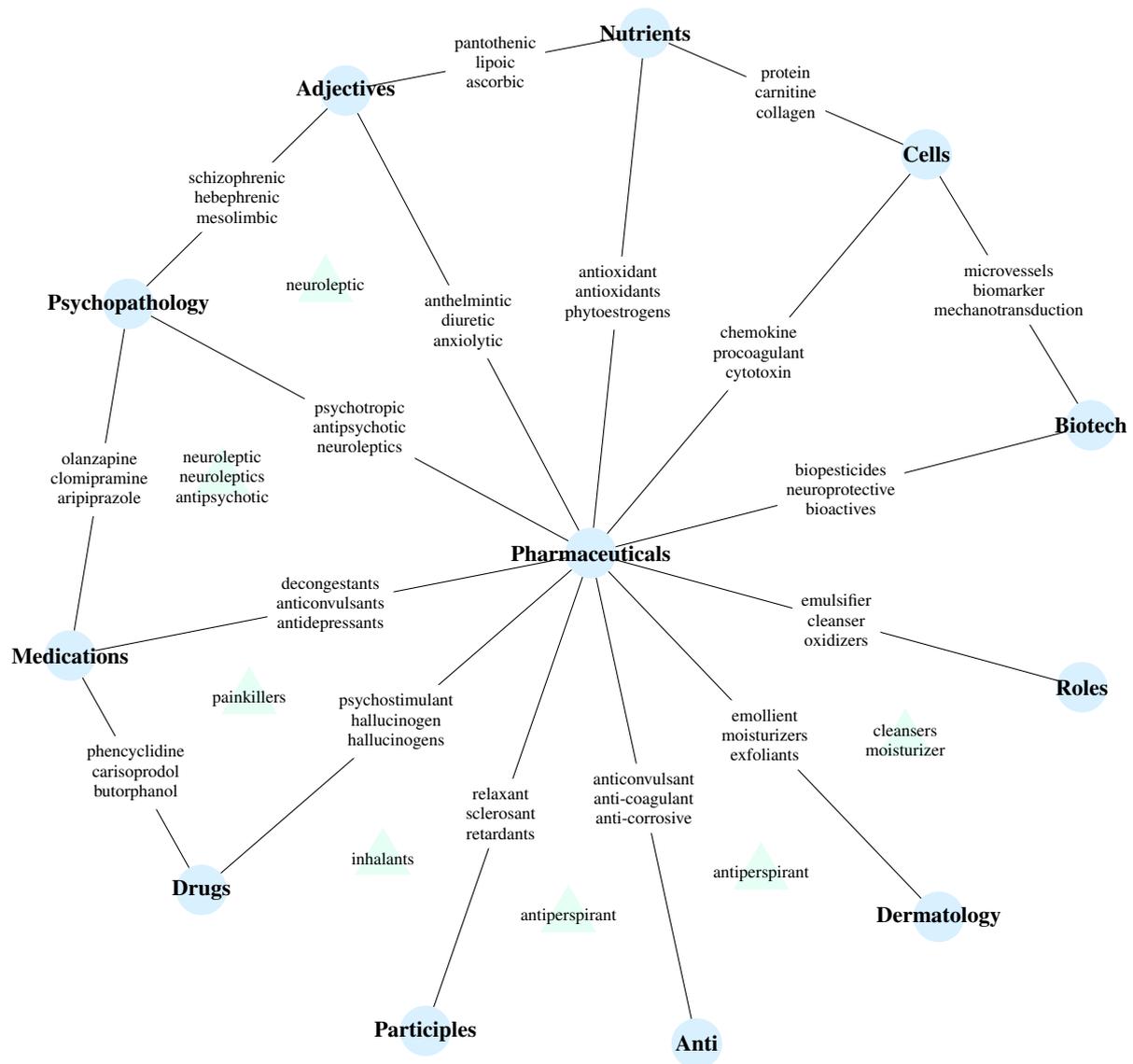

Various subgraphs of the graph of related components are shown in
Fig.~\ref{fig:grammar},
Fig.~\ref{fig:instruments}, and
Fig.~\ref{fig:pharmaceuticals}.
Additional graphs are presented in Appendix~\ref{app:moregraphs}.

The one-word names of all 512 components of the ICA-transformed hidden states of the Llama~3~70B model are presented in Appendix~\ref{app:components}.
By detailed analysis of the components, we can find there e.g. more than 20 components specifying different languages (Dutch, Japanese, Scandinavian, Italian, ...), more than 15 components specifying geographical objects (Rivers, Cities, Countries, Mountains, Islands, Places, ...),  and more than 10 components grouping words ending or beginning by specific letters. We can also find the main part-of-speech-tags (Nouns, Adjectives, Verbs, Pronouns, Prepositions, Adverbs). Not all categories are visible from the one-word component names. For this analysis, we used also longer component descriptions generated by GPT4o. The set of components observed always characterises the training data of the model. For example, in this case we found more than 40 components related to coding.\footnote{The numbers of components found are only approximate, some components might not be named well due to limited number of words used for naming or because GPT4o simply do not recognize the common feature of words and outputs something very general.}

The component graphs are quite similar across different models (for examples, see Appendix~\ref{app:compare}).

\fix{diskutovat vrstvy}

\fix{diskutovat 1024}

\fix{Jakým způsobem jsme vybrali ukázky.}

\todotm{počty sousedů}

\section{Conclusion}

We have presented an account of meaning as a set of semantic features,
which can be represented by a binary vector. We show how to estimate these
binary vectors from the hidden states of a LLM. Specifically, we show that hidden
states in Llama3 LLM represent semantic features, which can be found through
ICA. The resulting components are interpretable and can be combined as semantic
features. This is a promising interpretability technique. In future work,
it can be applied to e.g. differences between models, differences between layers in a single model,
analysis of specific textual domains, and other interpretability tasks.



\bibliography{main}

\appendix

\section{Prompts for Component Naming}
\label{app:gptprompt}

The following prompts were used with GPT4o to name the components.

System message: ``You are a Conceptual Grouping and Naming System, designed to analyze a given group of words and identify a common theme or characteristic.''

Prompt: ``Given the following group of words, provide a short name that
encapsulates what they have in common. If possible, use just one word.'' If the
last sentence of this prompt is omitted, the resulting names are generally more
specific, but substantially longer, which makes it hard to display them in a
graph.

\section{Similar Components from Different Models}
\label{app:compare}

\begin{figure*}

\vspace{-1cm}

 \hbox{}\hfill  \begin{tikzpicture}[scale=7,
compnode/.style={circle,fill,draw,blue!40!cyan!15,minimum size=20pt,inner sep=0pt},
innode/.style={regular polygon,regular polygon sides=3,green!40!cyan!10,fill,draw,minimum size=25pt,inner sep=0pt},
tedge/.style={fill=white,rounded corners,opacity=0.9,text opacity=1}]
      \draw[font=\small\bfseries]
        (-0.005, -0.004) node[compnode,label={[align=center]center:Music}] (191){}
        (-0.781, -0.667) node[compnode,label={[align=center]center:Specialists}] (144){}
        (0.878, 0.422) node[compnode,label={[align=center]center:Tools}] (174){}
        (0.996, 0.021) node[compnode,label={[align=center]center:Luminaries}] (401){}
        (-1.0, -0.211) node[compnode,label={[align=center]center:Enthusiasts}] (316){}
        (0.6, 0.757) node[compnode,label={[align=center]center:Indian}] (215){}
        (0.25, 0.941) node[compnode,label={[align=center]center:Functions}] (231){}
        (-0.962, 0.307) node[compnode,label={[align=center]center:Categories}] (372){}
        (-0.731, 0.682) node[compnode,label={[align=center]center:Abstracts}] (390){}
        (0.154, -0.998) node[compnode,label={[align=center]center:French}] (17){}
        (-0.4, 0.93) node[compnode,label={[align=center]center:Analysis}] (158){}
        (-0.365, -0.956) node[compnode,label={[align=center]center:Cinema}] (437){}
        (0.862, -0.506) node[compnode,label={[align=center]center:Historical}] (118){}
        (0.559, -0.829) node[compnode,label={[align=center]center:Actions}] (19){}
        (-0.669, -0.33) node[innode,label={[align=center,font=\scriptsize]center:drummers\\instrumentalists\\pianists}] (x144x316){}
        (0.577, 0.437) node[innode,label={[align=center,font=\scriptsize]center:veena}] (x174x215){}
        (0.696, -0.183) node[innode,label={[align=center,font=\scriptsize]center:purcell\\corelli}] (x401x118){}
        (-0.072, 0.716) node[innode,label={[align=center,font=\scriptsize]center:create\_song\\play\_music\\compose\_music}] (x231x158){}
        (-0.636, 0.37) node[innode,label={[align=center,font=\scriptsize]center:musics}] (x372x390){}
        (-0.081, -0.734) node[innode,label={[align=center,font=\scriptsize]center:opéra}] (x17x437){};
      \begin{scope}[-,align=center,font=\scriptsize, on background layer]
        \draw (191) to node[tedge,pos=0.3] {musics\\harmonies\\modulations} (390);
        \draw (191) to node[tedge,pos=0.6] {create\_melody\\create\_chord\\compose\_music} (231);
        \draw (191) to node[tedge,pos=0.55] {saxophone\\clarinet\\veena} (174);
        \draw (191) to node[tedge,pos=0.5] {trombonist\\saxophonist\\percussionist} (144);
        \draw (191) to node[tedge,pos=0.5] {debussy\\brahms\\beethoven} (401);
        \draw (191) to node[tedge,pos=0.6] {musique\\compositeur\\violoniste} (17);
        \draw (191) to node[tedge,pos=0.5] {composing\\improvising\\singing} (19);
        \draw (191) to node[tedge,pos=0.5] {music\\musician\\musik} (372);
        \draw (191) to node[tedge,pos=0.6] {operatic\\soundtrack\\musical} (437);
        \draw (191) to node[tedge,pos=0.5] {continuo\\ayres\\gasparini} (118);
        \draw (191) to node[tedge,pos=0.45] {sangeeta\\veena\\shankar} (215);
        \draw (191) to node[tedge,pos=0.5] {musicians\\instrumentalists\\percussionists} (316);
        \draw (191) to node[tedge,pos=0.5] {harmonized\_notes\\new\_song\\compose\_music} (158);
        \draw (144) to node[tedge] {mathematicians\\geoscientists\\astrophysicists} (316);
        \draw (144) to node[tedge] {screenwriter\\filmmaker\\dramatist} (437);
        \draw[white] (174) to node[] {} (401);
        \draw[white] (174) to node[] {} (215);
        \draw (401) to node[tedge] {molière\\marlowe\\moliere} (118);
        \draw (316) to node[tedge] {artists\\farmers\\students} (372);
        \draw[white] (215) to node[] {} (231);
        \draw (231) to node[tedge] {build\_house\\compose\_song\\solve\_problem} (158);
        \draw[white] (372) to node[] {} (390);
        \draw[white] (390) to node[] {} (158);
        \draw (17) to node[tedge] {comédie\\théâtre\\cinéma} (437);
        \draw[white] (17) to node[] {} (19);
        \draw[white] (118) to node[] {} (19);
      \end{scope}
    \end{tikzpicture}\hfill\hbox{}

\vspace{1.5cm}
    
 {\small
\centering
\begin{tabular}{p{.12\textwidth}>{\em}p{.327\textwidth} p{.12\textwidth}>{\em}p{.327\textwidth}}
\toprule
Component & Words & Component & Words \\
\midrule
Actions & tuned, serenaded, orchestrated, harmonized, modulated & Triple & quartet, quintet, trio, four-part, five-part \\
\bottomrule
\end{tabular}}

\vspace{.5cm}

 \caption{This is component number 191 from the ICA-transformed hidden states of the Llama~3~8B model. The 2 components that did not fit in the graph are listed in the table bellow. See the caption of Figure~\ref{fig:triangle} for explanation of the graphic symbols.}
 \label{fig:music}
 \end{figure*}
\begin{figure*}
 \hbox{}\hfill  \begin{tikzpicture}[scale=7,
compnode/.style={circle,fill,draw,blue!40!cyan!15,minimum size=20pt,inner sep=0pt},
innode/.style={regular polygon,regular polygon sides=3,green!40!cyan!10,fill,draw,minimum size=25pt,inner sep=0pt},
tedge/.style={fill=white,rounded corners,opacity=0.9,text opacity=1}]
      \draw[font=\small\bfseries]
        (0.002, -0.005) node[compnode,label={[align=center]center:Musicians}] (181){}
        (0.377, -0.939) node[compnode,label={[align=center]center:Specialists}] (721){}
        (0.979, -0.211) node[compnode,label={[align=center]center:Thinkers}] (26){}
        (0.765, -0.654) node[compnode,label={[align=center]center:Poetry}] (760){}
        (-0.167, -1.00) node[compnode,label={[align=center]center:DiverseProfessions}] (560){}
        (-0.892, -0.429) node[compnode,label={[align=center]center:Tools}] (612){}
        (-0.979, -0.008) node[compnode,label={[align=center]center:Breathing}] (551){}
        (-0.893, 0.412) node[compnode,label={[align=center]center:Teams}] (298){}
        (-0.58, -0.843) node[compnode,label={[align=center]center:Categories}] (450){}
        (0.344, 0.908) node[compnode,label={[align=center]center:Suffixes}] (241){}
        (-0.084, 0.979) node[compnode,label={[align=center]center:Loanwords}] (860){}
        (0.94, 0.296) node[compnode,label={[align=center]center:Russian}] (322){}
        (-0.57, 0.81) node[compnode,label={[align=center]center:Wiktionary}] (496){}
        (0.711, 0.664) node[compnode,label={[align=center]center:Traditions}] (684){}
        (0.094, -0.728) node[innode,label={[align=center,font=\scriptsize]center:instrumentalists\\bassists\\organists}] (x721x560){}
        (0.72, 0.031) node[innode,label={[align=center,font=\scriptsize]center:kabalevsky\\rachmaninoff\\tchaikovsky}] (x26x322){}
        (-0.245, 0.669) node[innode,label={[align=center,font=\scriptsize]center:komponist\\konzert\\jazzmusiker}] (x860x496){}
        (0.62, 0.359) node[innode,label={[align=center,font=\scriptsize]center:balalaika}] (x322x684){};
      \begin{scope}[-,align=center,font=\scriptsize, on background layer]
        \draw (181) to node[tedge,pos=0.5] {guitar\\piano\\guitars} (450);
        \draw (181) to node[tedge,pos=0.5] {prokofiev\\kabalevsky\\balalaika} (322);
        \draw (181) to node[tedge,pos=0.5] {shakuhachi\\mandolin\\mandolins} (612);
        \draw (181) to node[tedge,pos=0.5] {clarinets\\clarinet\\oboes} (551);
        \draw (181) to node[tedge,pos=0.5] {orchestra\\orchestras\\orquesta} (298);
        \draw (181) to node[tedge,pos=0.5] {shamisen\\balalaika\\didgeridoo} (684);
        \draw (181) to node[tedge,pos=0.5] {saxophonists\\violinists\\pianists} (560);
        \draw (181) to node[tedge,pos=0.5] {saxophonist\\pianist\\violinist} (721);
        \draw (181) to node[tedge,pos=0.5] {virtuosic\\symphonic\\melodic} (241);
        \draw (181) to node[tedge,pos=0.5] {klavier\\jazzmusiker\\schlagzeug} (496);
        \draw (181) to node[tedge,pos=0.5] {folksinger\\sång\\chorale} (760);
        \draw (181) to node[tedge,pos=0.5] {saint-saens\\liszt\\shostakovich} (26);
        \draw (181) to node[tedge,pos=0.5] {musiker\\koncert\\musik} (860);
        \draw (721) to node[tedge] {agronomists\\hydrologists\\biochemists} (560);
        \draw (721) to node[tedge] {lyricist\\poet\\lyricists} (760);
        \draw (26) to node[tedge] {tennyson\\mickiewicz\\lermontov} (760);
        \draw (26) to node[tedge] {lermontov\\chekhov\\dostoyevsky} (322);
        \draw (560) to node[tedge] {doctors\\lawyers\\farmers} (450);
        \draw (612) to node[tedge] {knife\\knives\\sword} (450);
        \draw[white] (612) to node[] {} (551);
        \draw[white] (551) to node[] {} (298);
        \draw (298) to node[tedge] {gemeinschaft\\verein\\verwaltungsgemeinschaft} (496);
        \draw[white] (241) to node[] {} (860);
        \draw[white] (241) to node[] {} (684);
        \draw (860) to node[tedge] {diskussion\\jazzmusiker\\komponist} (496);
        \draw[white] (322) to node[] {} (684);
      \end{scope}
    \end{tikzpicture}\hfill\hbox{}

\vspace{1cm}

 {\small
\centering
\begin{tabular}{p{.12\textwidth}>{\em}p{.327\textwidth} p{.12\textwidth}>{\em}p{.327\textwidth}}
\toprule
Component & Words & Component & Words \\
\midrule
Base & bassists, bassist, basslines, bassoons, bassline & Verbs & fiddle, fiddles, fiddling, fiddlers, fiddler \\
Concepts & tunings, improvisations, orchestrations, musiques, musics & Accents & guitare, guitariste, musique, musicien, pianiste \\
Genres & folk-rock, jazz, blues-rock, jazz-rock, folk-pop & Verbing & drumming, trumpeting, strumming, fingering, fiddling \\
URLs & vocals/guitar, vocalist/guitarist, singer/guitarist, actress/singer, guitar/vocals & Acoustic & 'music, \_music, musicians, music, sonor \\
Oracle & orchester, orchestral, orchestra, obo, orchest & Suffix & cornet, clarinet, clarinets, cornett, drumset \\
\bottomrule
\end{tabular}}
  \caption{This is component number 181 from the ICA-transformed hidden states of the Llama 3 70B model (with 1024 ICA components),
representing \emph{Musicians}. The outer circle shows the components that share words with this component.
The 10 components that did not fit in the graph are listed in the table bellow (together with top 5 words that combine
the listed component and the central component in the graph). See the caption of Figure~\ref{fig:triangle} for explanation of the
graphic symbols.
  }
 \label{fig:musicians}
 \end{figure*}

To illustrate how similar the components can be, we present the graphs of
related components for components centered on ``music'', obtained from three
different models. Compare the graph in Figure~\ref{fig:instruments}
(\emph{Instruments}, Llama 70B) with Figure ~\ref{fig:music} (\emph{Music},
Llama 8B) and Figure ~\ref{fig:musicians} (\emph{Musicians}, Llama 70B, ICA to
1024 components).

\section{More Component Graphs}
\label{app:moregraphs}

\begin{figure*}
 \hbox{}\hfill  \begin{tikzpicture}[scale=6.5,
compnode/.style={circle,fill,draw,blue!40!cyan!15,minimum size=20pt,inner sep=0pt},
innode/.style={regular polygon,regular polygon sides=3,green!40!cyan!10,fill,draw,minimum size=25pt,inner sep=0pt},
tedge/.style={fill=white,rounded corners,opacity=0.9,text opacity=1}]
      \draw[font=\small\bfseries]
        (0.005, -0.0) node[compnode,label={[align=center]center:Pregnancy}] (7){}
        (0.363, -0.942) node[compnode,label={[align=center]center:Sexuality}] (448){}
        (-0.137, 1.0) node[compnode,label={[align=center]center:Children}] (353){}
        (0.349, 0.95) node[compnode,label={[align=center]center:Pre-events}] (442){}
        (-0.893, -0.427) node[compnode,label={[align=center]center:Actions}] (486){}
        (-0.12, -1.0) node[compnode,label={[align=center]center:Inflammation}] (221){}
        (-0.982, -0.007) node[compnode,label={[align=center]center:Birthplaces}] (501){}
        (-0.592, 0.811) node[compnode,label={[align=center]center:Ages}] (272){}
        (0.982, 0.249) node[compnode,label={[align=center]center:Adjectives}] (138){}
        (0.986, -0.236) node[compnode,label={[align=center]center:Adjectives}] (193){}
        (0.752, 0.676) node[compnode,label={[align=center]center:Inter/Intra}] (74){}
        (-0.576, -0.82) node[compnode,label={[align=center]center:Surgeries}] (266){}
        (-0.9, 0.412) node[compnode,label={[align=center]center:Duration}] (311){}
        (0.762, -0.666) node[compnode,label={[align=center]center:Anatomy}] (466){}
        (0.423, -0.603) node[innode,label={[align=center,font=\scriptsize]center:oviduct\\uterine\\cervix}] (x448x466){};
      \begin{scope}[-,align=center,font=\scriptsize, on background layer]
        \draw (7) to node[tedge,pos=0.5] {insemination\\pregnancy\\uterus} (448);
        \draw (7) to node[tedge,pos=0.5] {baby\\babies\\infant} (353);
        \draw (7) to node[tedge,pos=0.5] {gestational\\obstetrical\\neonatal} (193);
        \draw (7) to node[tedge,pos=0.5] {fertilization\\implantation\\insemination} (486);
        \draw (7) to node[tedge,pos=0.5] {amniotic\\chorionic\\dichorionic} (138);
        \draw (7) to node[tedge,pos=0.5] {intrapartum\\intrauterine\\transplacental} (74);
        \draw (7) to node[tedge,pos=0.5] {c-section\\cesarean\\caesarean} (266);
        \draw (7) to node[tedge,pos=0.5] {2-month-old\\3-month-old\\two-month-old} (272);
        \draw (7) to node[tedge,pos=0.5] {periumbilical\\uterine\\uterus} (466);
        \draw (7) to node[tedge,pos=0.5] {birth\\født\\births} (501);
        \draw (7) to node[tedge,pos=0.5] {20-week\\nine-month\\16-week} (311);
        \draw (7) to node[tedge,pos=0.5] {pre-birth\\pre-term\\preborn} (442);
        \draw (7) to node[tedge,pos=0.5] {pre-eclampsia\\preeclampsia\\eclampsia} (221);
        \draw (448) to node[tedge] {vulvar\\testicles\\urethral} (466);
        \draw (448) to node[tedge] {dyspareunia\\menorrhagia\\priapism} (221);
        \draw (353) to node[tedge] {5-year-olds\\8-year-olds\\10-year-olds} (272);
        \draw (353) to node[tedge] {preadolescent\\pre-teens\\pre-teen} (442);
        \draw (442) to node[tedge] {pre-fight\\pre-race\\pre-conference} (74);
        \draw (486) to node[tedge] {enucleation\\cauterization\\intubation} (266);
        \draw[white] (486) to node[] {} (501);
        \draw (221) to node[tedge] {pericardiocentesis\\cholesteatoma\\aneurysm} (266);
        \draw[white] (501) to node[] {} (311);
        \draw (272) to node[tedge] {38-year\\34-year\\36-year} (311);
        \draw (138) to node[tedge] {talismanic\\algorithmic\\paradigmatic} (193);
        \draw (138) to node[tedge] {transoceanic\\intertrochanteric\\subtrochanteric} (74);
        \draw (193) to node[tedge] {ureteral\\pharyngeal\\conjunctival} (466);
      \end{scope}
    \end{tikzpicture}\hfill\hbox{}

\vspace{1cm}
    
 {\small
\centering
\begin{tabular}{p{.15\textwidth}>{\em}p{.25\textwidth} p{.15\textwidth}>{\em}p{.25\textwidth}}
\toprule
Component & Words & Component & Words \\
\midrule
Anti & anti-abortion, antepartum, anti-choice, antenatal, pro-life & Latin & partum, antepartum, peripartum, gravidarum, intrapartum \\
Suffixes & dystocia, hydramnios, pre-eclampsia, eclampsia, polyhydramnios & Hyphenated & pre-natal, new-born, pre-term, post-partum, pre-mature \\
\bottomrule
\end{tabular}}
\caption{This is component number 7 from the ICA-transformed hidden states of the Llama 3 70B model,
representing \emph{Roles}. The outer circle shows the components that share words with this component.
The 4 components that did not fit in the graph are listed in the table bellow (together with top 5 words that combine
the listed component and the central component in the graph). See the caption of Figure~\ref{fig:triangle} for explanation of the
graphic symbols.}
 \label{fig:pregnancy}
 \end{figure*}
\begin{figure*}

\vspace{0cm}

 \hbox{}\hfill  \begin{tikzpicture}[scale=5.75,
compnode/.style={circle,fill,draw,blue!40!cyan!15,minimum size=20pt,inner sep=0pt},
innode/.style={regular polygon,regular polygon sides=3,green!40!cyan!10,fill,draw,minimum size=25pt,inner sep=0pt},
tedge/.style={fill=white,rounded corners,opacity=0.9,text opacity=1}]
      \draw[font=\small\bfseries]
        (0.003, -0.006) node[compnode,label={[align=center]center:Roles}] (99){}
        (-0.244, -0.987) node[compnode,label={[align=center]center:Groups}] (291){}
        (-0.681, -0.748) node[compnode,label={[align=center]center:Components}] (321){}
        (0.984, 0.173) node[compnode,label={[align=center]center:Crimes}] (186){}
        (-0.009, 1.0) node[compnode,label={[align=center]center:Transformation}] (176){}
        (0.25, -0.988) node[compnode,label={[align=center]center:Specialists}] (510){}
        (0.808, 0.57) node[compnode,label={[align=center]center:Tools}] (239){}
        (0.49, 0.866) node[compnode,label={[align=center]center:British}] (198){}
        (-0.973, 0.126) node[compnode,label={[align=center]center:Intensifiers}] (219){}
        (-0.947, -0.308) node[compnode,label={[align=center]center:Grammar}] (63){}
        (0.685, -0.755) node[compnode,label={[align=center]center:Football}] (287){}
        (0.956, -0.338) node[compnode,label={[align=center]center:Hyphenates}] (16){}
        (-0.506, 0.851) node[compnode,label={[align=center]center:Programming}] (33){}
        (-0.815, 0.543) node[compnode,label={[align=center]center:Sounds}] (65){}
        (-0.346, -0.652) node[innode,label={[align=center,font=\scriptsize]center:makers}] (x291x321){}
        (0.003, -0.742) node[innode,label={[align=center,font=\scriptsize]center:engravers\\animators}] (x291x510){}
        (-0.61, -0.397) node[innode,label={[align=center,font=\scriptsize]center:modifier}] (x321x63){}
        (-0.192, 0.693) node[innode,label={[align=center,font=\scriptsize]center:serializer}] (x176x33){};
      \begin{scope}[-,align=center,font=\scriptsize, on background layer]
        \draw (99) to node[tedge,pos=0.5] {maker\\reader\\trainer} (321);
        \draw (99) to node[tedge,pos=0.5] {updater\\formatter\\serializer} (33);
        \draw (99) to node[tedge,pos=0.5] {answerers\\throwers\\talkers} (291);
        \draw (99) to node[tedge,pos=0.5] {tackler\\goalscorer\\defender} (287);
        \draw (99) to node[tedge,pos=0.5] {shuffler\\screamer\\wobbler} (65);
        \draw (99) to node[tedge,pos=0.5] {analyser\\optimiser\\stabiliser} (198);
        \draw (99) to node[tedge,pos=0.5] {magnifier\\eraser\\grater} (239);
        \draw (99) to node[tedge,pos=0.5] {mobilizer\\normalizer\\visualizer} (176);
        \draw (99) to node[tedge,pos=0.5] {problem-solver\\problem-solvers\\record-breaker} (16);
        \draw (99) to node[tedge,pos=0.5] {stealer\\kidnapper\\harasser} (186);
        \draw (99) to node[tedge,pos=0.5] {brightener\\softener\\sharpener} (219);
        \draw (99) to node[tedge,pos=0.5] {landscaper\\preparer\\embroiderer} (510);
        \draw (99) to node[tedge,pos=0.5] {determiner\\modifier\\intensifier} (63);
        \draw (291) to node[tedge] {players\\leaders\\speakers} (321);
        \draw (291) to node[tedge] {calligraphers\\agronomists\\mathematicians} (510);
        \draw (321) to node[tedge] {usage\\form\\forms} (63);
        \draw (186) to node[tedge] {law-breaking\\warmongering\\fear-mongering} (16);
        \draw[white] (186) to node[] {} (239);
        \draw (176) to node[tedge] {initializes\\initialize\\serialize} (33);
        \draw (176) to node[tedge] {formalised\\politicised\\industrialised} (198);
        \draw (510) to node[tedge] {futbolista\\ \rub футболіст \rue \\futebolista} (287);
        \draw[white] (239) to node[] {} (198);
        \draw[white] (219) to node[] {} (63);
        \draw[white] (219) to node[] {} (65);
        \draw (287) to node[tedge] {cup-winning\\goalscoring\\game-winner} (16);
        \draw[white] (33) to node[] {} (65);
      \end{scope}
    \end{tikzpicture}\hfill\hbox{}

\vspace{0.33cm}
    
 {\small
\centering
\begin{tabular}{p{.12\textwidth}>{\em}p{.327\textwidth} p{.12\textwidth}>{\em}p{.327\textwidth}}
\toprule
Component & Words & Component & Words \\
\midrule
Users & adopter, admirer, purchaser, hearer, drinker & Finance & acquirers, depositor, acquirer, issuer, lender \\
Artisan & engraver, embroiderer, carvers, engravers, etcher & Emotive & stunner, eye-catcher, crowd-pleaser, pleaser, shocker \\
Insults & tossers, fucker, waster, stinker, poser & Sports & vaulter, paddler, skater, grappler, skier \\
Rework & rebuilder, resizer, re-animator, rewriter, reloader & Temperatures & heaters, scorcher, heater, cooker, igniter \\
Acronyms & forger, inverter, topper, forager, inker & Retail & wholesaler, discounters, merchandiser, discounter, retailer \\
Pharmaceuticals & emulsifier, vasoconstrictor, vasodilator, cleanser, oxidizers & Derision & encourager, basher, bashers, accuser, hater \\
Variables & messagesender, menuprovider, cashapelayer, filereader, javamailsender & Connected & combiner, unifier, connecter, connector, coupler \\
Apparel & jumper, blazer, choker, jumpers, fascinator & Animals & grouper, adder, adders, snapper, bee-eater \\
Pre-events & pre-accelerator, preloader, pre-processor, preconditioner, preprocessors & Nonsense & streeter, soother, screener, seeder, peeler \\
MachineLearning & recommender, annotator, labelbinarizer, featureextractor, gradientboostingregressor & Writing & transcriber, writer, scribbler, corrector, rewriter \\
Buildings & renovator, remodeler, constructor, renovators, remodelers & Instruments & synthesizer, sequencer, arranger, composer, improviser \\
Message & messager, communicator, sender, talker, transmitter & Hardware & multiplexer, multiplexers, demodulator, inverters, demultiplexer \\
Mathematics & annihilator, expanders, minimizers, normalizer, minimizer & Misspellings & reciever, processer, controler, processers, convertor \\
\bottomrule
\end{tabular}}
\caption{This is component number 99 from the ICA-transformed hidden states of the Llama 3 70B model,
representing \emph{Roles}. The outer circle shows the components that share words with this component.
The 26 components that did not fit in the graph are listed in the table bellow (together with top 5 words that combine
the listed component and the central component in the graph). See the caption of Figure~\ref{fig:triangle} for explanation of the
graphic symbols.}
 \label{fig:roles}
 \end{figure*}

To further illustrate the landscape of semantic features covered by the ICA
components, we present the graphs centered on the following components from the Llama~3~70B model:

\begin{itemize}
    \item \emph{Pregnancy} in Figure~\ref{fig:pregnancy},
    \item \emph{Roles} in Figure~\ref{fig:roles}.
\end{itemize}

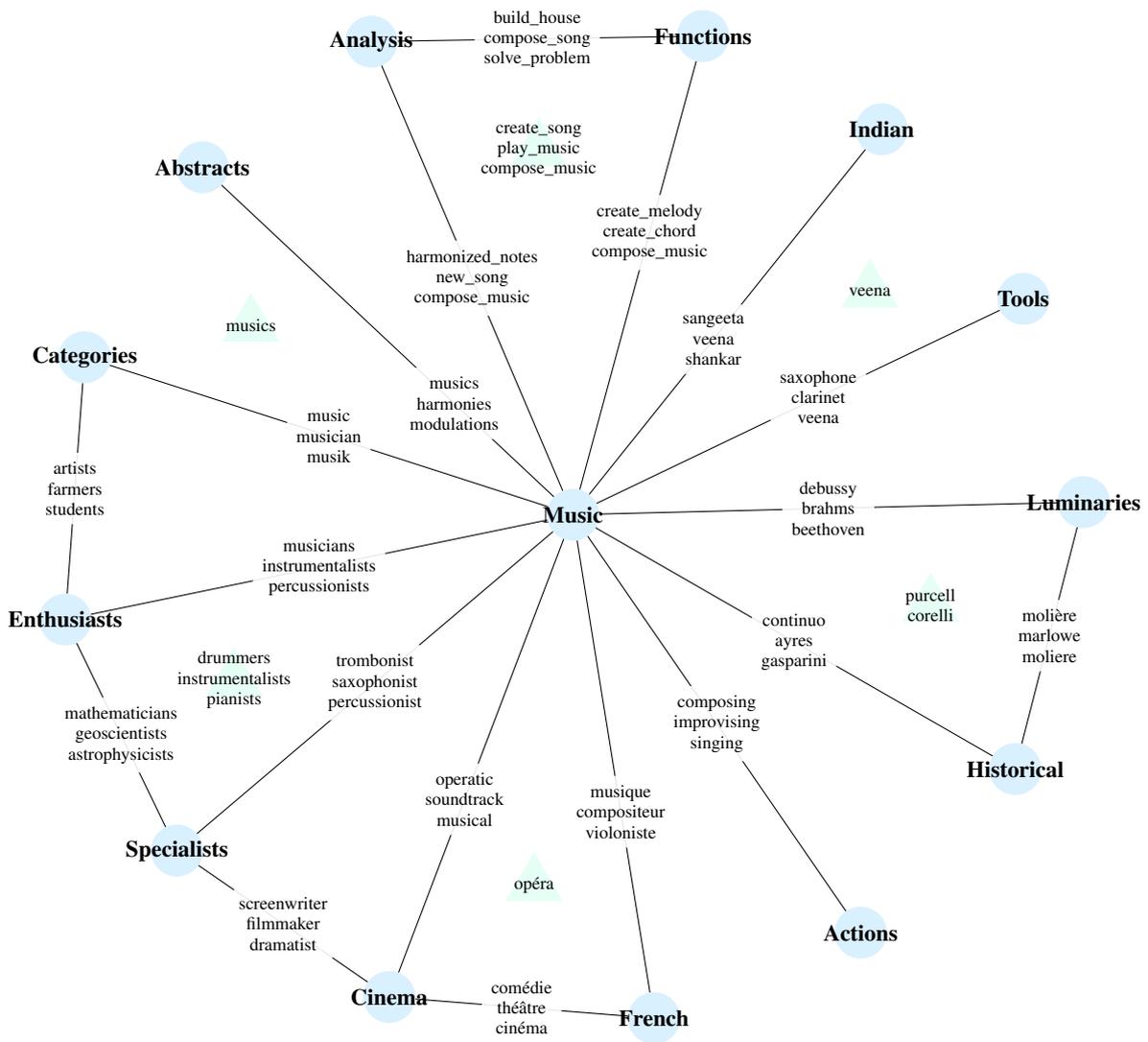
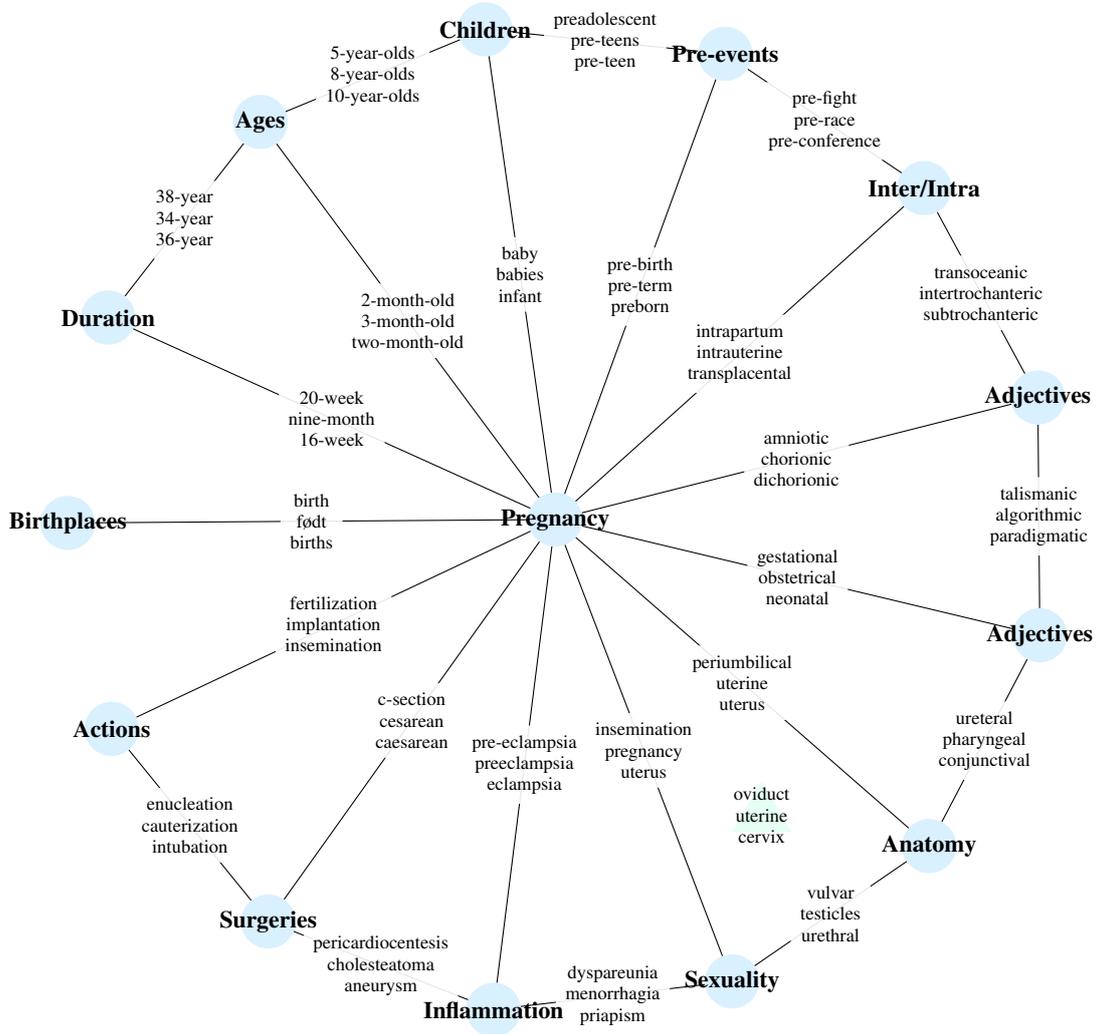
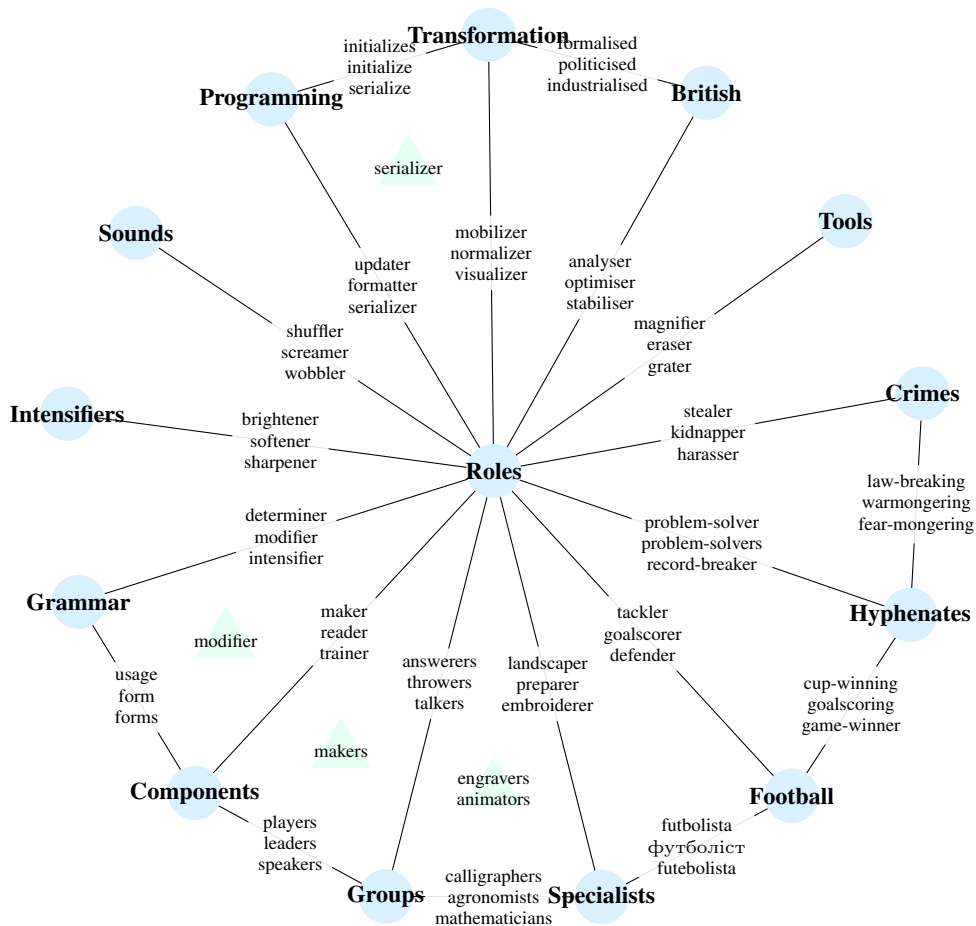
\onecolumn
\section{Names of the Components}
These are the names of all 512 components from the ICA-transformed hidden states of the Llama~3~70B model.
\label{app:components}
\small
\begin{longtable}{rccccc}
0 & Brands & Handwashing & Configuration & Ceremony & Names \\
5 & Rivers & JavaScript & Pregnancy & Files & Fitness \\
10 & Superficial & Seasons & Tribes & Church & Abbreviations \\
15 & Actions & Hyphenates & Surnames & Film & Names \\
20 & Places & Obscure & Scores & \rub Информатика \rue & Suffixes \\
25 & Shapes & Bakedgoods & Users & Slavic & Wikipedia \\
30 & Esoteric & Decimals & Classrooms & Programming & Scale \\
35 & Places & Codes & Instruments & Collaborative & Latin \\
40 & Uncertainty & Styles & Academia & Exponents & Wordle \\
45 & Latin & Arabic & Modules & Verses & Sciences \\
50 & Currency & Views & Times & Prefixes & Indigenous \\
55 & Botanicals & Numbers & Fractions & Seven & Neighborhoods \\
60 & Languages & Variables & Brands & Grammar & Pharmaceuticals \\
65 & Sounds & Apparel & Containers & Scales & Taxa \\
70 & Typo & Compound & Islands & Verbs & Inter/Intra \\
75 & Microsoft & Locations & Repetition & Artisan & Syndromes \\
80 & Abbreviations & Hebrew & Minerals & Pairs & Celebrities \\
85 & Anesthetics & Places & Common & Surnames & Towns \\
90 & Writers & Nonsense & Polish & Times & Compound \\
95 & Places & PlaceNames & Slang & Enzymes & Roles \\
100 & Limitations & E-Services & Phonetics & Abbreviations & JavaScript \\
105 & Variants & DataTypes & Append & Multi & Controversies \\
110 & Mathematics & Languages & Years & Pixels & Light \\
115 & Occupations & Prefixes & Years & Digital & Fields \\
120 & French & Taxa & Musicians & Verbs & Status \\
125 & Numbers & Color & Absence & Message & Complexity \\
130 & JSON & Places & Birds & Greek & LaTeX \\
135 & Non- & Love & Identifiers & Adjectives & Squared \\
140 & Pandas & Temporal & Numbers & Municipalities & Controllers \\
145 & German & Phonemes & Instruments & Acronyms & Abbreviations \\
150 & Hardware & Mythical & Lexicon & Gastronomy & Adverbs \\
155 & Strings & Destinations & Dutch & Variables & Animals \\
160 & Common & Items & Primes & Mountains & Verbs \\
165 & Websites & Pagination & Suffix & Surnames & Fish \\
170 & Ownership & Decades & Numbers & Existence & Electrochemical \\
175 & Literature & Transformation & Routes & Scottish & Decimals \\
180 & Drugs & 2020s & Employment & Excess & Towns \\
185 & Scandinavian & Crimes & Numbers & Scrabble & Suffixes \\
190 & Gaming & Medications & Functional Groups & Adjectives & Defeat \\
195 & Airlines & Names & Documentos & British & Years \\
200 & Names & Names & Numbers & Variables & Acronyms \\
205 & Substantivos & Cyrillic & Hawaiian & HTTP & Pairs \\
210 & Cells & Places & Politics & Names & Topics \\
215 & South African & Double Letters & Abbreviations & Cities & Intensifiers \\
220 & Surnames & Inflammation & Elements & Derision & Adjectives \\
225 & Superlatives & Oid & Nutrients & Countries & Polysyllabic \\
230 & Versions & Pathogens & Libraries & TechSolutions & Unresolved \\
235 & Abstracts & Decimals & Apostrophes & Exceptional & Tools \\
240 & Subdivisions & Insults & Nonsense & Services & Fields \\
245 & Dates & Hardware & Virtues & Nicknames & Land \\
250 & Historical & Aviation & Housing & Towns & Angles \\
255 & Inclusive & Clubs & Fractions & Airlines & Thriving \\
260 & Suffix & Truncations & Establishments & Paths & Emotive \\
265 & Testing & Surgeries & Years & Brands & Noise \\
270 & Writing & Parameters & Ages & Identifiers & Lines \\
275 & Brands & Prepositions & Units & Equal. & Lists \\
280 & Keywords & Vowels & Suffixes & Trees & Fruits \\
285 & Suffixes & Ordinal & Football & Names & Historical \\
290 & JavaLibraries & Groups & Connected & Healthcare & Context \\
295 & ClothingCleanliness & Surnames & Numbers & Variables & Biblical \\
300 & Decimals & Times & Variables & Functions & Accents \\
305 & Organization & Prefixes & Vietnamese & Hyphenated & Prefixes \\
310 & CompoundWords & Duration & Anti & Input & Hyphenates \\
315 & Corporations & Acronyms & Streams & Demonyms & Qualities \\
320 & Multilingual & Components & Biotech & Russian & Websites \\
325 & Chess & Numbers & Exceptions & Charity & Self \\
330 & Sorrow & Neighborhoods & Binary & Vowels & Acronyms \\
335 & Boolean & Past & Headers & Adjectives & Prefixes \\
340 & Numbers & Finance & Retail & Suffixes & Cyrillic \\
345 & Size & Abbreviations & Suffixy & Names & Automotive \\
350 & Temperature & Data & Hyphenated & Children & Redundancy \\
355 & Years & Maritime & Numbers & Mixed & News \\
360 & Dogs & Misspellings & Places & Distances & Negative \\
365 & Cooked & Template & Frameworks & Verbs & M-Names \\
370 & Initials & Utilities & Towns & Females & Diverse \\
375 & Nailcare & Family & Numbers & Palindrome & Dates \\
380 & Seeds & Astonishment & Storeys & Time & Furniture \\
385 & Affluence & Fractions & Japanese & UIComponents & Numbers \\
390 & Components & Psychopathology & Sports & Invertebrates & Traditions \\
395 & Negative & Abbreviations & Misspellings & Pronouns & Botany \\
400 & Luxury & Hyphenated & Image & Conferences & Names \\
405 & MathScience & Yoruba & Rework & Concatenation & Compounds \\
410 & Italian & Surnames & Royalty & Particles & Olde \\
415 & Years & Coordinates & Abbreviations & Military & Regions \\
420 & Anatomy & Identifiers & Origins & Transformation & Indian \\
425 & Keywords & Suffixes & Hygiene & ZIP Codes & Verblish \\
430 & Latin & Acronyms & Authentication & -isms & Places \\
435 & Mountains & UIComponents & Surnames & HTTP Status Codes & Acronyms \\
440 & Creation & Human-Centric & Pre-events & OnlineServices & Verbs \\
445 & Times & Buildings & Abbreviations & Sexuality & Places \\
450 & DatabaseConnection & Invented & Numbers & Lengths & Wine \\
455 & Abbreviations & Repository & JavaScript & Livestock & Numbers \\
460 & User & Verbing & Simile & Adjectives & Numbers \\
465 & Times & Anatomy & Models & Films & Participles \\
470 & Gaelic & Families & Getters & Materials & Actions \\
475 & Indonesian & Dermatology & Prices & Chinese & Actionable \\
480 & Adjectives & Portmanteaus & MachineLearning & Distances & Directions \\
485 & K-words & Actions & Occupations & Ingredients & Concepts \\
490 & Australia & Related & Years & Surnames & Gerunds \\
495 & CSS Selectors & Measurements & Cancers & Places & Legal \\
500 & Impact & Birthplaces & Northeast & Based & Greetings \\
505 & Numbers & Lesions & Abbreviations & Abbreviations & Attributes \\
510 & Specialists & Temperatures &  &  &  \\

\end{longtable}
\twocolumn

\end{document}